# COPE: Chain-Of-Thought Prediction Engine for Open-Source Large Language Model Based Stroke Outcome Prediction from Clinical Notes

Yongkai Liu, PhD[1], Helena Feng[1], Bin Jiang, MD[1], Yixin Wang, MS[1], Max Wintermark, MD[2], David S. Liebeskind, MD[3], Michael Moseley, PhD[1], Maarten Lansberg, MD[4], Gregory Albers, MD[4], Jeremy Heit, MD, PhD[1], Greg Zaharchuk, MD, PhD[1]

[1]Department of Radiology, Stanford University, Stanford, CA, USA
[2]Department of Neuroradiology, University of Texas MD Anderson Center, Houston, TX, USA
[3]Department of Neurology, UCLA, Los Angeles, CA, USA
[4]Department of Neurology, Stanford, Stanford, CA, USA

# ABSTRACT

Predicting outcomes in acute ischemic stroke (AIS) guides clinical decision-making, patient counseling, and resource allocation. Clinical notes contain rich contextual information, but their unstructured nature limits their use in traditional predictive models. We developed and evaluated the Chain-of-Thought (CoT) Outcome Prediction Engine (COPE), a reasoning-enhanced large language model framework, for predicting 90-day functional outcomes after AIS from unstructured clinical notes. This study included 464 AIS patients with discharge summaries and 90-day modified Rankin Scale (mRS) scores. COPE uses a two-step CoT framework based on sequential open-source LLaMA-3-8B models: the first generates clinical reasoning, and the second outputs an mRS prediction. We compared COPE with GPT-4.1, ClinicalBERT, a structured variable-based machine learning model (Clinical ML), and a single-step LLM without CoT. Performance was evaluated using mean absolute error (MAE), accuracy within +/-1 mRS point, and exact accuracy. COPE achieved an MAE of 1.01 (95% CI 0.92-1.11), +/-1 accuracy of 74.4% (69.9, 78.8%), and exact accuracy of 32.8% (28.0, 37.6%), comparable to GPT-4.1 and superior to ClinicalBERT [MAE 1.24 (1.13-1.36)], Clinical ML [1.28 (1.18-1.39)], and the single-step LLM [1.20 (1.09-1.33)]. Subgroup analyses showed consistent performance across sex and age, with slightly higher error among older patients, those undergoing thrombectomy, and those with longer summaries. These findings demonstrate that COPE, a lightweight, interpretable, and privacy-preserving open-source framework, provides an accurate and practical solution for outcome prediction from unstructured clinical text.

# INTRODUCTION

Predicting patient outcome is an important part of the clinical decision-making process, helpful for patient expectations and resource management[1]. Free-text clinical notes, such as discharge summaries, offer comprehensive documentation of a patient's hospital stay, encompassing clinical narratives, treatments, laboratory and imaging results, and follow-up plans. Despite their rich contextual and linguistic detail, these documents remain underutilized in predictive modeling[2,3]. One clinical domain where the utilization of free-text clinical notes holds particular promise is stroke, which is a leading cause of death globally. Many survivors endure significant functional impairments, which can greatly diminish their quality of life[4]. Given the high frequency and morbidity of acute ischemic stroke (AIS) coupled with a standardized, well-accepted, and patient-centric outcome scoring system (the 90-day modified Rankin Scale [mRS])[3-5], AIS outcome prediction is an ideal test case for evaluating methods to predict outcome from free-text clinical notes[5].

Conventional approaches to outcome prediction often require manual extraction of structured clinical variables from unstructured discharge summaries—a time-consuming process that must be performed before applying machine learning (ML) algorithms[6,7]. These methods are not only resource-intensive but also risk omitting nuanced clinical context embedded in free-text narratives. Recent advances in natural language processing (NLP), including language modeling techniques, have enabled more automated extraction of information from clinical text, improving stroke prediction performance. For example, Sung et al.[5] used a domain-adapted transformer model—Bio+Discharge Summary Bidirectional Encoder Representations from Transformers (BERT)—to predict functional outcomes after ischemic stroke, achieving comparable or superior accuracy to traditional structured-data models. However, many of these approaches rely on supervised fine-tuning, which requires large volumes of annotated clinical data that are often unavailable due to privacy, access, and cost constraints[8].

Large language models (LLMs) offer a potential solution by enabling zero-shot or few-shot inference directly from raw clinical text[9]. While proprietary models such as GPT-4 have demonstrated promise in healthcare applications, their adoption is limited by concerns around data privacy, cost, and lack of adaptability[10,11]. In contrast, open-source LLMs provide transparent, customizable, and cost-effective alternatives. Despite these advantages, the application of open-source LLMs in clinical outcome prediction remains largely unexplored[12,13].

To address this gap, we introduce the Chain-of-Thought (CoT) Outcome Prediction Engine (COPE)—a zero-shot, reasoning-enhanced framework built on the open-source LLaMA-3-8B model—and present stroke outcome prediction as a demonstrative use case. COPE employs a two-step architecture that uses CoT prompting to first generate intermediate clinical reasoning and then produce a structured outcome prediction. This design enhances interpretability and output reliability compared with standard prompting approaches, which often struggle with rigid output formats or limited clinical context. By leveraging open-source LLMs, COPE provides a transparent, scalable, and privacy-preserving alternative to proprietary systems. We evaluate its performance against a state-of-the-art proprietary model (GPT-4.1), a conventional language model (ClinicalBERT), a traditional clinical variable–based machine learning model (Clinical ML), and a single-step LLM without reasoning—demonstrating that open-source, CoT-enhanced models can match or exceed the performance of closed systems while reducing the need for manual feature engineering and enabling zero-shot prediction without fine-tuning.

**METHODS**

*Data Collection and Preparation*

The dataset consisted of 464 patients with AIS, enrolled from a single institutional registry at Stanford University Hospital. All patients were evaluated for thrombectomy eligibility between 2010 and 2023. For each patient, the discharge summary—an unstructured clinical document capturing the patient's hospital course, diagnostics, interventions, and discharge plan—was used as the primary input for modeling. Long-term functional outcomes were assessed using the mRS, an ordinal patient-centered measure of post-stroke disability that ranges from 0 (no symptoms) to 6 (death)[14]. The mRS score was recorded at a median of 90 days following discharge (range: 60–120 days). Patients who died during the initial hospitalization were excluded, as death information can be directly extracted from the discharge note, representing an observed outcome rather than a predictive task that could bias model evaluation. A flowchart detailing the inclusion and exclusion process is shown in Figure 1. Prior to analysis, all discharge summaries were de-identified by removing protected health information (PHI), in accordance with HIPAA guidelines.

*Chain-of-thought Outcome Prediction Engine (COPE)*

*Model Structure*   Figure 2 illustrates the overall framework of COPE and provides an example of the two-step process. COPE is a two-step pipeline built with two off-the-shelf, open-source

LLaMA-3-8B-Instruct models[15] run purely in inference mode; no additional fine-tuning or weight updates were performed. The first model, referred to as the "reasoning LLM," receives the entire discharge summary and produces a structured chain-of-thought output containing key clinical cues, provisional reasoning, and a tentative prognosis. The second model, the "extraction LLM," receives this reasoning text and returns a single mRS score between 0 to 6. By decoupling free-form clinical reasoning from final score generation, COPE not only enforces strict output formatting but also improves interpretability and predictive performance.

*Prompt Design and Optimization*   The patient dataset was randomly divided into a 20% prompt-exploration subset, used to optimize prompts for both LLMs, and an 80% held-out test set, stratified by mRS scores. The final prompts were selected based on predictive accuracy and the validity of numeric outputs on the exploration subset and were subsequently evaluated on the independent test set. The final prompt texts and associated details are provided in the Supplementary Table 1.

**Ablation Studies and Comparison Models**

To evaluate the effectiveness of this framework, we conducted a series of ablation and comparative studies, including comparisons with GPT-4.1, ClinicalBERT, and a clinical variable–based machine learning model (Clinical ML). To isolate the contribution of reasoning, we also compared COPE against a single-step LLM without reasoning.

*ClinicalBERT Baseline:* We fine-tuned the publicly available Bio_ClinicalBERT model[16] specifically for the mRS score prediction task. Since discharge summaries often exceed the 512-token input limit of BERT-based models, each summary was split into overlapping 512-token segments with a 50-token overlap. Each segment was treated as an independent instance during training. During inference, we applied the same chunking strategy and aggregated the model's predictions across segments using the method described by Huang et al.[3] for scalable readmission prediction.

*Clinical ML Baseline:* To benchmark against traditional machine learning approaches, we implemented a Support Vector Regression model trained on structured features manually extracted from discharge summaries. These variables included demographic information (age, sex), prior medical history (stroke, hypertension, diabetes, and atrial fibrillation), transfer status, clinical scores (baseline, 24-hour, and discharge National Institutes of Health Stroke Scale [NIHSS]), relevant laboratory results (HbA1c and LDL), treatment indicators (e.g., intravenous tissue plasminogen activator [IV tPA] administration and EVT), procedural and hospital course information (Thrombolysis in Cerebral Infarction [TICI] score and presence of procedure-related complications), and discharge destination.

*GPT-4.1:* The closed-source GPT-4.1 model was evaluated using the same optimized prompt and two-step chain-of-thought structure developed for the COPE framework. The first step generated intermediate clinical reasoning from the discharge summary, followed by a second step that extracted the final mRS using a structured parsing prompt. This design allows for a controlled comparison by holding the reasoning process and prompt engineering constant. The model was deployed in a secure, in-hospital environment at Stanford University Hospital, maintained by Stanford Medicine's Technology and Digital Solutions team, in compliance with institutional policies for handling protected health information.

Single-step LLM (no reasoning): In this baseline, the same open-source LLM used in COPE directly predicted the mRS from the discharge summary without generating intermediate reasoning, allowing evaluation of whether explicit reasoning improves performance beyond a straightforward end-to-end prediction. Prompt exploration was performed using the same exploration subset as COPE to guide the LLM toward strong performance while ensuring it directly output a single integer mRS score in the correct format.

*Performance Evaluation*

To assess model performance in predicting outcome, we employed three metrics: mean absolute error (MAE), exact accuracy (ACC), and ±1 accuracy (±1 ACC). MAE measures the average absolute difference between the predicted and actual 90-day mRS scores. ACC represents the percentage of exact matches between predicted and true scores, while ±1 ACC reflects the percentage of predictions falling within ±1 point of the ground truth—providing a tolerance-aware evaluation appropriate for ordinal scales.

*Subgroup and Sensitivity Analysis Design*

To assess model robustness and potential performance disparities of COPE, subgroup analyses were conducted across demographic, clinical, and documentation-related factors. Evaluation was stratified by sex (male vs. female), treatment status (endovascular thrombectomy [EVT] vs. non-EVT), discharge summary length (quartile-based word count), and patient age group (<46, 46–64, 65–80, and >80 years). For each subgroup, model performance was quantified using MAE with 95% confidence intervals derived via bootstrap resampling. Sensitivity analyses were performed to examine whether model error patterns remained consistent across strata and to identify potential sources of bias in prediction accuracy.

*Statistical Analysis*

To evaluate statistical significance across performance metrics, several tests were employed. A bootstrap-based hypothesis test[17] was used to assess differences in MAE, ACC, and ±1 ACC. P-values less than 0.05 were considered statistically significant. To correct for multiple comparisons, the Benjamini–Hochberg procedure was applied, controlling the false discovery rate at 0.05. Bootstrap resampling was also used to estimate 95 % confidence intervals for all metrics.

# RESULTS

*Patient Characteristics*

A total of 1,462 patients with acute ischemic stroke were initially identified from the Stanford University Hospital registry. After applying exclusion criteria—specifically, missing 90-day mRS data (n = 937) and in-hospital death (n = 61)—a final cohort of 464 patients (median age, 70 years; IQR, 60–80 years; 254 male patients) was included in the study. Demographic characteristics are summarized in Table 1. The dataset was then randomly divided into two subsets: 20% (n = 92) for prompt exploration and 80% (n = 372) for model testing.

*Performance Analysis*

Table 2 presents a comparative evaluation of five models for stroke outcome prediction: GPT-4.1, Clinical BERT, Clinical ML, the Single-Step LLM, and the proposed COPE framework.

The COPE model achieved an MAE of 1.01 (95% CI, 0.92–1.11), ±1 ACC of 74.4% (69.9–78.8%), and exact ACC of 32.8% (28.0–37.6%). These results were statistically indistinguishable from those of GPT-4.1 [MAE = 1.00 (0.90–1.09); ±1 ACC = 77.9% (73.7–82.0%); ACC = 32.5% (28.0–37.4%); $p$ = 0.72, 0.11, and 0.96, respectively].

Compared with COPE, Clinical BERT demonstrated higher error and lower accuracy [MAE = 1.24 (1.13–1.36); ±1 ACC = 64.5% (59.7–69.4); ACC = 28.5% (23.9–33.1); $p$ < 0.001 for MAE and ±1 ACC; p = 0.13 for ACC]. Clinical ML also underperformed [MAE = 1.28 (1.18–1.39); ±1 ACC = 60.5% (55.6–65.6); ACC = 26.9% (22.3–31.5); p < 0.001 for MAE and ±1 ACC; p = 0.06 for ACC]. Finally, the Single-Step LLM baseline performed worse than COPE [MAE = 1.20 (1.09–1.33); ±1 ACC = 67.5% (62.9–72.0); ACC = 29.7% (25.4–34.0); p < 0.001 for MAE and ±1 ACC; p = 0.09 for ACC].

**Subgroup and Sensitivity Analyses of COPE**

Subgroup analysis examined model error across demographic, clinical, and documentation strata, including sex, treatment status, note length, and age (Figure 3). The difference between males (MAE, 0.99; 95% CI, 0.86–1.12; n = 206) and females (1.04; 0.90–1.18; n = 166) was minimal. By treatment status, non-EVT cases showed lower error (0.80; 0.62–0.99; n = 84) compared with EVT cases (1.07; 0.97–1.18; n = 288). Error increased with note length across quartiles: Q1 <1958 words (0.87; 0.69–1.05), Q2 1958–2389 words (1.03; 0.86–1.21), Q3 2390–2843 words (1.06; 0.86–1.27), and Q4 ≥2843 words (1.09; 0.90–1.28). By age, MAE was 1.03 (0.70–1.40; n = 30) for <46 years, 0.90 (0.76–1.05; n = 106) for 46–64 years, 0.91 (0.77–1.07; n = 153) for 65–80 years, and 1.32 (1.11–1.55; n = 83) for >80 years, indicating the highest error among the oldest patients.

**DISCUSSION**

We developed and evaluated COPE, a CoT reasoning–enhanced LLM framework for predicting patient outcomes from free-text clinical notes, demonstrating its performance for stroke outcomes based on discharge summaries. Outcome prediction is a fundamentally important task, with benefits to patients, clinicians, resource managers, and researchers, and methods to perform this using free-text, unstructured clinical notes remain underexplored.

The open-source COPE approach demonstrated performance comparable to a state-of-the-art proprietary LLM, GPT-4.1, across multiple outcome prediction metrics. Remarkably, this was achieved using LLaMA 3–8B—a smaller, open-source model with approximately 60–100× fewer parameters and substantially lower computational requirements compared to GPT-4.1. This result highlights the potential of task-specific design—such as incorporating chain-of-thought reasoning—to unlock high performance from lightweight models. In addition to matching GPT-4.1 in predictive accuracy, COPE offers practical advantages in transparency, reproducibility, and clinical deployability. Open-source architectures like LLaMA 3 enable greater flexibility for fine-tuning, domain adaptation, and integration into privacy-sensitive healthcare workflows, making COPE an accessible and scalable solution for real-world clinical applications.

Unlike traditional single-step prompting, COPE separates the reasoning process into two sequential stages: first generating an intermediate clinical rationale, then producing the final prediction. This two-step design consistently outperformed the single-step baseline, indicating that prompting the model to reason explicitly through the clinical context before prediction enhances both accuracy and robustness. During development, we observed that single-step LLMs often produced free-text or mixed-format outputs when given detailed clinical context or reasoning instructions, even when explicitly instructed to return only an integer between 0 and 6. Only highly constrained prompts reliably yielded valid outputs. Paradoxically, more elaborate prompts, which often improve reasoning in other domains, degraded performance in this structured clinical prediction setting. These findings underscore a fundamental limitation of single-step LLMs: they struggle to balance complex reasoning with rigid output constraints. By decoupling reasoning from prediction, the COPE framework addresses this challenge, producing more reliable, interpretable, and clinically actionable predictions.

Previous approaches to outcome prediction have primarily relied on clinical variable–based machine learning models[18–20]. In comparison, our LLM-based framework achieved significantly lower error and higher predictive accuracy. This improvement likely stems from the model's ability to process rich, unstructured clinical narratives, which provide contextual cues and nuanced insights beyond what structured data alone can offer. Traditional models depend on manually extracted clinical features—a time-consuming and labor-intensive process that limits scalability. In contrast, our approach leverages open-source, pretrained LLMs to eliminate the need for manual feature engineering while capturing complex relationships and contextual dependencies within unstructured text.

We found that COPE's prediction errors were concentrated in subgroups characterized by higher clinical complexity and greater documentation burden. The higher error observed among EVT cases likely reflects case-mix and outcome heterogeneity: patients undergoing EVT are typically more severely affected (higher baseline NIHSS), experience variable reperfusion success, and are at risk for peri-procedural complications—all of which broaden the range of possible outcomes and increase prediction difficulty. Similarly, model performance decreased with longer discharge notes, which often include procedural templates, radiology addenda, and multi-service narratives that may dilute or obscure salient clinical cues-reflecting the broader challenge of adapting LLMs to the noise and complexity of real-world clinical data[21]. Finally, the higher error among patients older than 80 years may reflect wider variability in recovery trajectories, frailty, comorbidities, and discharge disposition, all of which likely contribute to less predictable functional outcomes.

Our study has several limitations. First, the models were evaluated using data from a single center, which may limit generalizability to other healthcare settings with different patient populations, clinical workflows, or documentation styles. However, the dataset encompasses diverse clinical presentations and real-world discharge documentation, supporting its relevance as a proof of concept. Future multicenter validation will be essential to confirm external generalizability and evaluate performance across heterogeneous practice environments. Second, the quality of prompt engineering likely influenced how the large language model made predictions, playing a critical role in optimizing stroke outcome estimation. In this study, prompt design was largely manual and guided by iterative trial and error, reflecting the current limitations of prompt standardization in clinical applications. Developing systematic or automated prompt optimization approaches may help reduce reliance on manual tuning and improve consistency in model reasoning. Lastly, the study did not assess clinician interaction with the model or its interpretability in clinical decision-making. Evaluating usability and trust among care providers will be critical for future translation.

In conclusion, we developed and evaluated COPE, a reasoning-enhanced large language model framework built on LLaMA 3–8B, which demonstrated strong performance in predicting 90-day stroke outcomes from free-text clinical discharge summaries. By integrating a two-step chain-of-thought architecture, COPE achieved performance comparable to the state-of-the-art proprietary model GPT-4.1 while remaining lightweight, interpretable, and privacy-preserving. These findings underscore the potential of open-source, reasoning-based LLMs to enable transparent and scalable clinical outcome prediction from unstructured text.

**ETHICS APPROVAL**

This study was conducted in accordance with the United States Health Insurance Portability and Accountability Act (HIPAA) of 1996 and was approved by the local Institutional Review Board (IRB-77549). In compliance with local IRB guidelines, as this is a retrospective study using de-identified clinical notes, the requirement for patient consent was waived by the Institutional Review Board.

# TABLES

**Table 1:** Summary of demographics for all acute ischemic stroke (AIS) patients (n=464) included in the study.

|  |  | Prompt Exploration Set (n=92) | Testing Set (n=372) |
|---|---|---|---|
| Male | | 48 (52.2) | 206 (55.4) |
| Age, yrs<br>Median (IQR) | | 71 (60, 81) | 70 (60, 79) |
| Hypertension | | 60 (65.2)<br>4.3% | 243 (65.3)<br>0.5% |
| Diabetes | | 27 (29.3) | 88 (23.7)<br>1.1% |
| Baseline NIHSS<br>Median (IQR) | | 15 (8, 20) | 14 (9, 20)<br>0.5% |
| iv-tPA | | 35 (38.0)<br>4.3% | 177 (47.6)<br>2.9% |
| EVT | | 76 (82.6) | 294 (79.0) |
| mRS scores at 90 days [3] | 0 | 10 (10.9) | 42 (11.3) |
| | 1 | 14 (15.2) | 56 (15.1) |
| | 2 | 11 (12.0) | 45 (12.1) |
| | 3 | 21 (22.8) | 83 (22.3) |
| | 4 | 19 (20.7) | 75 (20.2) |
| | 5 | 7 (7.6) | 30 (8.1) |
| | 6 | 10 (10.9) | 41 (11.0) |

Unless otherwise stated, data are presented as the number of patients and their corresponding percentage of the total. mRS: modified Rankin Scale; NIHSS: National Institutes of Health Stroke Scale; iv-tPA: intravenous tissue plasminogen activator; EVT: endovascular therapy. Percentages shown in gray indicate missing data for the variables below. If no data are missing, no percentage will be reported.

**Table 2:** Performance of different models for stroke outcome prediction. Comparison includes a state-of-the-art proprietary LLM (GPT-4.1), a conventional language model (Clinical BERT), a classical machine learning model (Clinical ML), and the proposed Chain-of-thought Outcome Prediction Engine (COPE). Performance metrics include mean absolute error (MAE), accuracy within ±1 mRS point (±1 ACC), and exact accuracy (ACC).

| Model Type | MAE | ±1 ACC (%) | ACC (%) |
|---|---|---|---|
| GPT 4.1 | 1.00 (0.90, 1.09) p = 0.72 | 77.9 (73.7, 82.0) p = 0.11 | 32.5 (28.0, 37.4) p = 0.96 |
| Clinical BERT | 1.24 (1.13, 1.36) p < 0.001 | 64.5 (59.7, 69.4) p < 0.001 | 28.5 (23.9, 33.1) p = 0.13 |
| Clinical ML | 1.28 (1.18, 1.39) p < 0.001 | 60.5 (55.6, 65.6) p < 0.001 | 26.9 (22.3, 31.5) p = 0.06 |
| Single-Step LLM | 1.20 (1.09, 1.33) p < 0.001 | 67.5 (62.9, 72.0) p < 0.001 | 29.7 (25.4, 34.0) p = 0.09 |
| **Chain-of-Thought Outcome Prediction Engine (COPE)** | **1.01 (0.92, 1.11)** | **74.4 (69.9, 78.8)** | **32.8 (28.0, 37.6)** |

Abbreviations: MAE, mean absolute error; ±1 ACC, prediction accuracy within ±1 mRS score; ACC, exact prediction accuracy of the mRS score; BERT, Bidirectional Encoder Representations from Transformers; Lower MAE and higher accuracy values indicate better performance. Values in parentheses indicate 95% confidence intervals. p-values were calculated using bootstrap resampling to assess statistical significance relative to the COPE model.

**Figures**

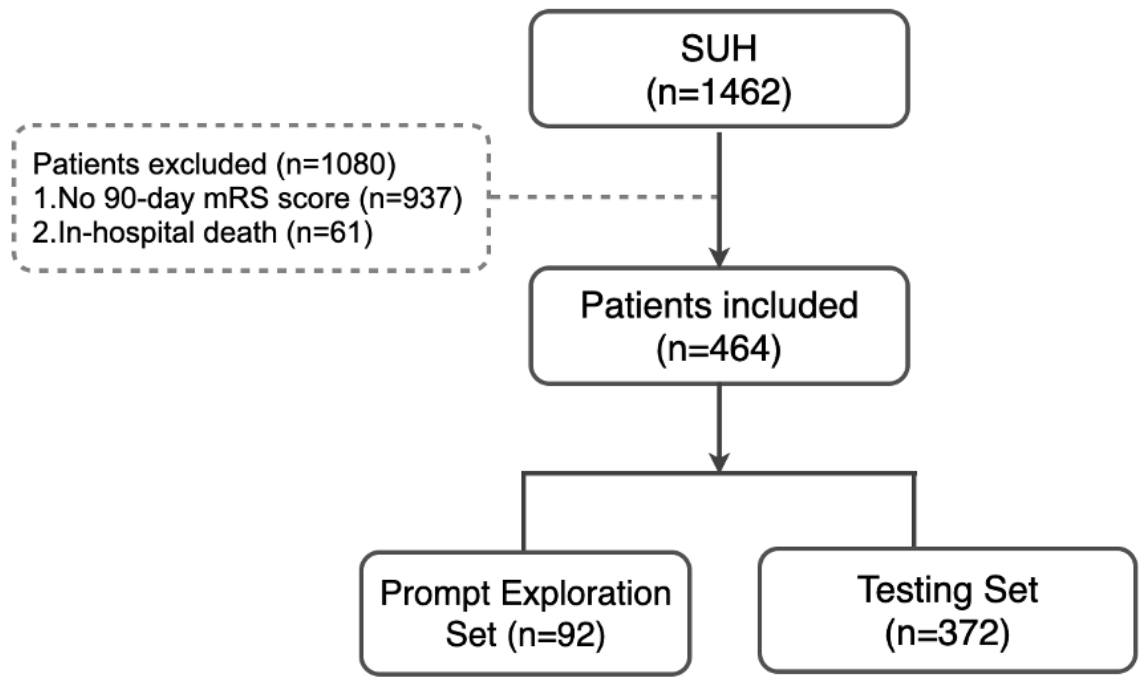

**Figure 1:** Flowchart showing patient inclusion and exclusion criteria. Patients were excluded if they lacked a 90-day modified Rankin Scale (mRS) score (n = 937) or experienced in-hospital death (n = 61). The remaining 464 patients were included and stratified by their mRS scores, then randomly divided into a prompt exploration set (20%) and a testing set (80%).

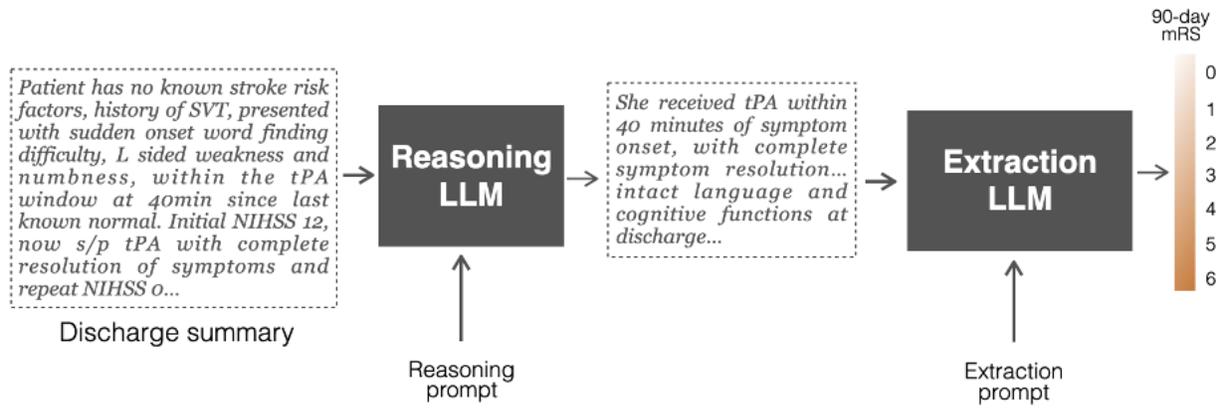

**Figure 2:** Overview of Chain-of-Thought Outcome Prediction Engine (COPE). The discharge summary was first fed into an LLM, utilizing a reasoning prompt to support the outcome prediction task. The output of the first LLM was then fed to the framework's second LLM step, which used an extraction prompt to derive a single mRS score from the reasoning provided by the first step.

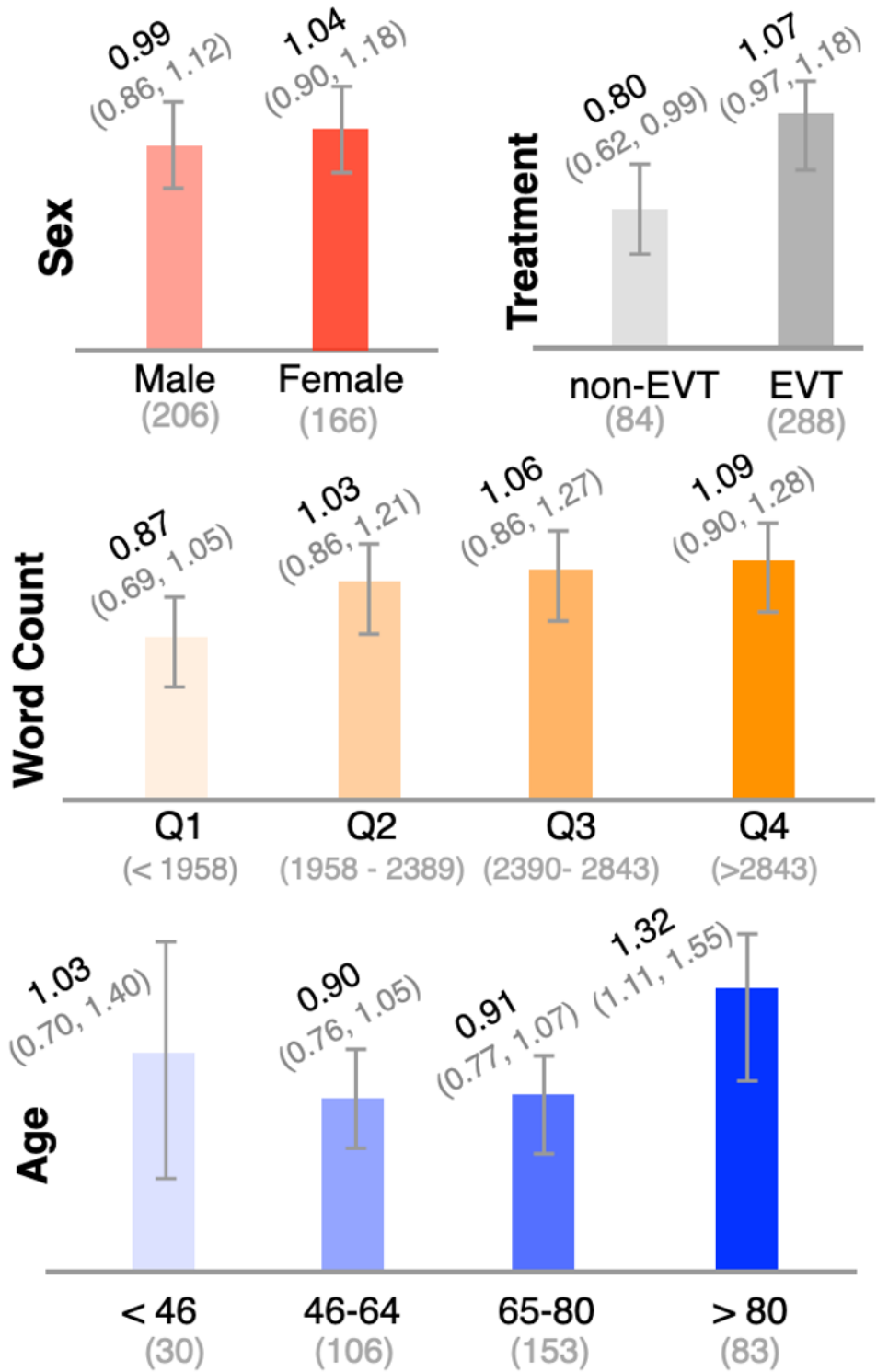

**Figure 3:** Mean absolute error (MAE) and 95% confidence intervals are shown across subgroups defined by sex, treatment status, documentation length (quartiles of note word count), and patient

age. Model error was similar between male and female patients but higher among those who underwent endovascular thrombectomy (EVT) compared with non-EVT cases. Prediction error increased with longer documentation length and was highest among patients older than 80 years, suggesting that clinical complexity and documentation burden contribute to reduced model accuracy.

# SUPPLEMENTARY MATERIALS

**Supplementary Table 1:** Prompts for Chain-of-thought Outcome Prediction Engine

| | |
|---|---|
| Reasoning Prompt | You are a highly experienced vascular neurologist at a tertiary comprehensive stroke center. You have been provided with a patient's discharge summary, which includes their clinical history, physical examination findings, imaging results, treatments administered, complications, and condition at discharge. Your task is to carefully analyze this summary and predict the patient's modified Rankin Scale (mRS) score at 90 days post-discharge.<br><br>The modified Rankin Scale (mRS) measures the degree of disability or dependence after a stroke, where:<br>0 = No symptoms<br>1 = No significant disability (able to carry out all usual activities)<br>2 = Slight disability (unable to perform all previous activities but can handle own affairs without assistance)<br>3 = Moderate disability (requires some help but can walk unassisted)<br>4 = Moderately severe disability (unable to walk without assistance and unable to attend to bodily needs without help)<br>5 = Severe disability (bedridden, incontinent, requiring constant nursing care)<br>6 = Dead<br><br>Please consider all relevant clinical details provided, including stroke severity, neurological deficits at discharge, rehabilitation potential, comorbidities, and social support. Make an evidence-based prediction of the single most likely mRS score at 90 days. If the information is incomplete or uncertain, use your best clinical judgment to select the most appropriate score. Provide a brief, clear rationale explaining how you arrived at this prediction, referencing key points from the discharge summary. |
| Extraction Prompt | Using the previous analysis results provided below, determine the most likely single modified Rankin Scale (mRS) score. If the analysis provides a range (e.g., '1 or 2'), select the value that reflects the more likely outcome based on the description. The output must be strictly a single integer between 0 and 6. Do not include any additional text or explanation. |